\documentclass[letterpaper, 10 pt, conference]{ieeeconf} 

\IEEEoverridecommandlockouts                              % This command is only needed if 

\usepackage[draft]{todonotes}
\usepackage{array}
\usepackage{tabularx}
\usepackage{graphicx}
\usepackage[caption=false]{subfig}

\title{\LARGE \bf Towards An Architecture-Centric Approach to Manage Variability of Cloud Robotics}

\author{Lei Zhang$^{1}$, Huaxi (Yulin) Zhang$^{2}$, Zheng Fang$^{3}$, Xianbo Xiang$^{4}$, Marianne Huchard$^{5}$ and Ren\'e Zapata$^{6}$
\thanks{*This work was supported by the National Natural Science Foundation of China under Grant No. 61300020, the Scientific Research Funds for Introduced Talents of Northeastern University under Grant No. 28720524.}
\thanks{$^{1}$Lei Zhang is with State Key Laboratory of Synthetical Automation for Process Industries, 
Northeastern University, Shenyang, China,
 {\tt\small zl.org.cn@gmail.com}}
\thanks{$^{2}$Huaxi (Yulin) Zhang is with MIS and INSSET,
Universit\'e de Picardie Jules Verne, 
33, rue Saint Leu, 80039 Amiens, France,
{\tt\small yulin.zhang@u-picardie.fr}}
\thanks{$^{3}$Zheng Fang is with State Key Laboratory of Synthetical Automation for Process Industries, 
Northeastern University, Shenyang, China,
 {\tt\small fangzheng@mail.neu.edu.cn}}
\thanks{$^{4}$Xianbo Xiang is with School of Naval Architecture and Ocean Engineering, Huazhong University of Science and Technology, 1037, Luoyu Road, 430074, Wuhan, China,
 {\tt\small xbxiang@hust.edu.cn}}
\thanks{$^{5}$Marianne Huchard is with LIRMM, UMR 5506,
CNRS et Universit\'e Montpellier 2,
161 rue Ada, 34392 Montpellier, France,
 {\tt\small huchard@lirmm.fr}}
\thanks{$^{6}$Ren\'e Zapata is with LIRMM, UMR 5506,
CNRS et Universit\'e Montpellier 2,
161 rue Ada, 34392 Montpellier, France,
 {\tt\small zapata@lirmm.fr}}
}

\begin{document}

\maketitle
\thispagestyle{empty}
\pagestyle{empty}
\maketitle

\begin{abstract}

Cloud robotics is a field of robotics that attempts to invoke Cloud technologies such as Cloud computing, Cloud storage, and other Internet technologies centered around the benefits of converged infrastructure and shared services for robotics. In a few short years, Cloud robotics as a newly emerged field has already received much research and industrial attention. 
The use of the Cloud for robotics and automation brings some potential benefits largely ameliorating the performance of robotic systems. However, there are also some challenges. First of all, from the viewpoint of architecture, how to model and describe the architectures of Cloud robotic systems? How to manage the variability of Cloud robotic systems? How to maximize the reuse of their architectures? 
In this paper, we present an architecture approach to easily design and understand Cloud robotic systems and manage their variability. 

\end{abstract}

\IEEEpeerreviewmaketitle

\section{Introduction}

Cloud robotics is a field of robotics that attempts to invoke Cloud technologies such as Cloud computing, Cloud storage, and other Internet technologies centered around the benefits of converged infrastructure and shared services for robotics~\cite{Kuffner2010}.
“Cloud Robotics” was firstly introduced by James Kuffner~\cite{Kuffner2010}. 
In a few short years, Cloud robotics as a newly emerged field has already received much research and industrial attention.

The use of Cloud computing for robotics and automation brings some potential benefits largely ameliorating the performance of robotic systems. Due to the limited capacities of on-board processing, storage and battery capacities, robotic devices are constrained to numerous limitations.
It not only solves the problems of robotic systems, such as on-board computation and storage limitation, asynchronization communication, compatibility problem of multi-robot systems\cite{Mohanarajah2014}, 
but also makes possibility of different directions or enhances their performance, such as remote brain, big data and shared knowledge-base, collective learning and intelligent behavior\cite{Qureshi2014}.

However, beyond these advantages, Cloud robotics also brings us many challenges. For example, from the view of architectures, how to construct the architectures of Cloud robotic systems? How to model these architectures? How to deploy these architectures in Clouds? How to reuse these architectures? How to manage the variability of these architectures?

In this paper, we propose a domain specific language -- CRALA trying to response the above questions. Our main contributions are to propose:
\begin{itemize}
\item an architecture-centric design process for Cloud robotic systems,
\item a domain specific language for architecture-centric Cloud robotic systems named CRALA.
\end{itemize}

The rest of the paper is organized as follows: We begin with an introduction of related concepts, background and related work of Architecture-centric Cloud robotics. We then present an overview of the architecture-centric design process for Cloud robotic systems. Then we describe the metamodel of CRALA with examples and how CRALA manages the variability of Cloud robotic systems. Afterwards, we present the implementation of CRALA. Finally, we finish with a discussion and future work.

\section{Background and Related work}

\subsection{Related concepts}~\label{sec:rc}
Architecture-centric Cloud robotics is a methodology of developing robotics systems on Clouds using architecture-centric development techniques. 

\textbf{Cloud computing}
Cloud computing is defined by the National Institute of Standards and Technology (NIST) as: "Cloud computing as a model for enabling ubiquitous, convenient, on-demand network access to a shared pool of configurable computing resources (e.g., networks, servers, storage, applications, and services) that can be rapidly provisioned and released with minimal management effort or service provider interaction~\cite{Mell2011}". 

Clouds offer services
that can be grouped into three categories: software as a service
(SaaS), platform as a service (PaaS), and infrastructure
as a service (IaaS)~\cite{Zhang2010b}.

\begin{enumerate}
\item Infrastructure as a Service: IaaS refers to on-demand
provisioning of infrastructural resources, usually in terms
of VMs (Virtual Machines). The cloud owner who offers IaaS is called an
IaaS provider.
\item Platform as a Service: PaaS refers to providing platform
layer resources, including operating system support and
software development frameworks.
\item Software as a Service: SaaS refers to providing on-demand
applications over the Internet.
\end{enumerate}

An explicit architecture of Cloud robotic system should cover these three design services in its architecture.  

\textbf{System/Software architectures.}
Traditionally, software architecture is a collection of models that capture a
software system’s principal design decisions in the form of
components (foci of system computation and data management),
connectors (foci of component interaction), and configurations
(specific arrangements of components and connectors intended to
solve specific problems) \cite{Perry1992}.
Generally speaking, a software system architecture~\cite{Taylor2009} gathers design decisions of the system. As the development of computer science, nowadays, a system is much more complex than before such as with the integration of "Internet of Things", "Cloud Computing" and "Robotics" etc.

\textbf{Architectures Modeling language.}
Architecture models are often expressed using A\textsc{dl}s (Architecture Description Language) that, in most cases, provides information
on the structure of the software system listing the components/services and connectors that the system is composed of. A system architecture could cover different abstraction levels, such as specification, configuration and assembly \cite{Zhang2010} and from different viewpoints \cite{Clements2002}. 
For Cloud robotic system, architecture model also needs to capture Cloud and robot design decisions.

\subsection{Related Work}
The description of Cloud robotic systems should cover robot description, web services/component description and cloud robotic system global architecture description. 

\textbf{Robot description.}
Robot description languages provide models of a robot and then design and implemented software components that work on the model components rather than the particular robot instance.  

The representative example of robot description language is the Unified Robot Description Format (URDF)~\cite{URDF}, which can be used to specify the kinematics and dynamics, the visual representation and the collision model of a robot. However, URDF is not designed for specifying robot components such as sensors, actuators, and control programs. 

COLLADA~\cite{Diankov2011} is an XML Schema designed for describing 3D objects including their kinematics. It mainly focuses on modeling information about scenes, geometry, physics, animations, and effects. But similar to URDF, it lacks elements for describing sensors, actuators and software.
SRDL~\cite{Kunze2011} focuses on modeling robot components, i.e. sensors, actuators and control programs, especially via capabilities to actions. 

Many work try to develop an OWL (Web Ontology Language) ontology to describe robots, such as ~\cite{Schlenoff2005,Chatterjee2005}
in specific domains or~\cite{Compton2009, Compton2009a} focusing on sensor ontology.

\textbf{Web service description.}
Web service description in robotics often serves to match the capabilities with robot components, such as PHOSPHORUS~\cite{Gil2001}, Larks~\cite{Sycara2002}, OWL-S~\cite{Martin2007} and SRDL\cite{Kunze2011}. 
In general, the term capability matchmaking refers to the process of matching an advertisement of a capability with a request. 

\textbf{Cloud Robotic description.}
All above work cover a part description of Cloud robotic systems, referring to robot description or web service description. However, it misses a language that fully covers all necessary aspects of Cloud robotic architecture, including the description of Clouds, robots and components/web services. 

\section{Architecture Design for Cloud robotics}

\begin{figure}[t]
\centering
\includegraphics[width=3in]{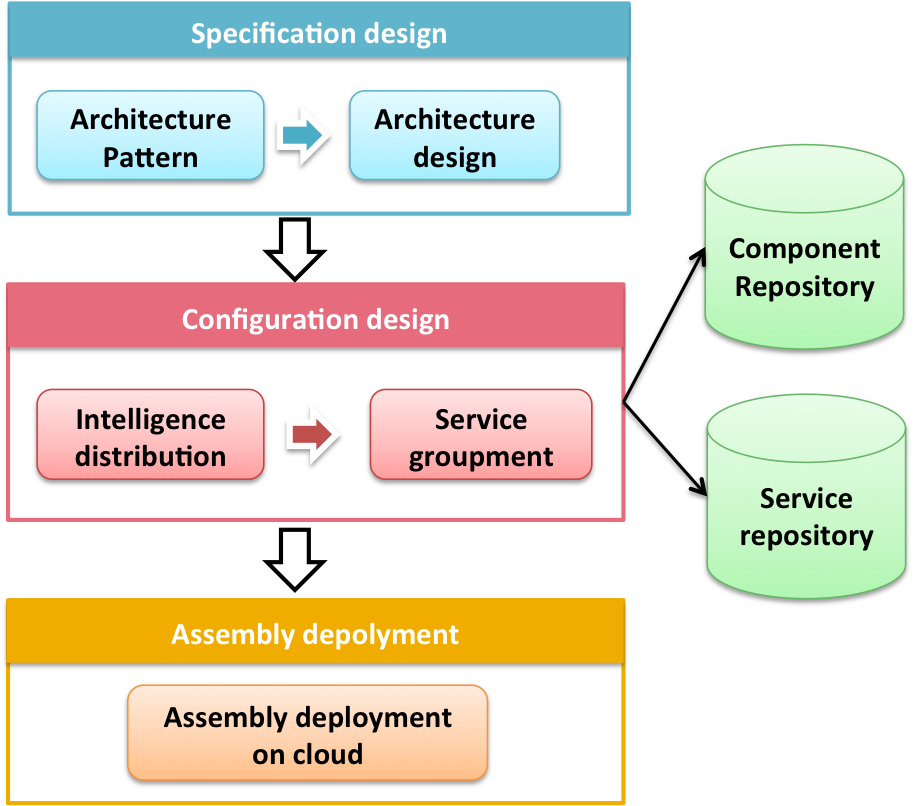}
\caption{Cloud robotic system architecture design}
\label{fig_process}
\end{figure}

The architecture design for Cloud robotic system is different from traditional software design, as it concerns two special aspects: Cloud-based systems and robotic systems. 
We identify the architecture-centric development process for Cloud robotic systems into three main phases, as shown in Figure~\ref{fig_process}. 

\begin{enumerate}
\item \textit{Specification design.} Architect or robotics engineers should choose a robotic architecture pattern for the system according the models of robots (hardware) and its functional tasks (objective), for example, a pioneer robot with a task of path planning. 
\item \textit{Configuration design.} 
\begin{itemize}
\item First of all, architect should consider how to distribute intelligence among robots and Cloud. That means, which components should be placed on the robot itself and which services should be placed on Cloud. How to choose the appropriate components or services from component or service repository. This design decision refers to different factors, including robot capacities, system non-functional properties such as real-time, security etc. 
\item Secondly, architect should choose operating system for their services, as in robotics domain, there exists some operating systems that are widely used, such as ROS\cite{ros}. Then, how to distribute these services in different virtual machines.
\end{itemize}

\item \textit{Assembly deployment.} Lastly, how to deploy this architecture model in Clouds, automatically or not? How to reflect and supervise a runtime model to prevent VMs failure etc.?
\end{enumerate}

\begin{figure}[t]
\centering
\includegraphics[width=0.4\textwidth]{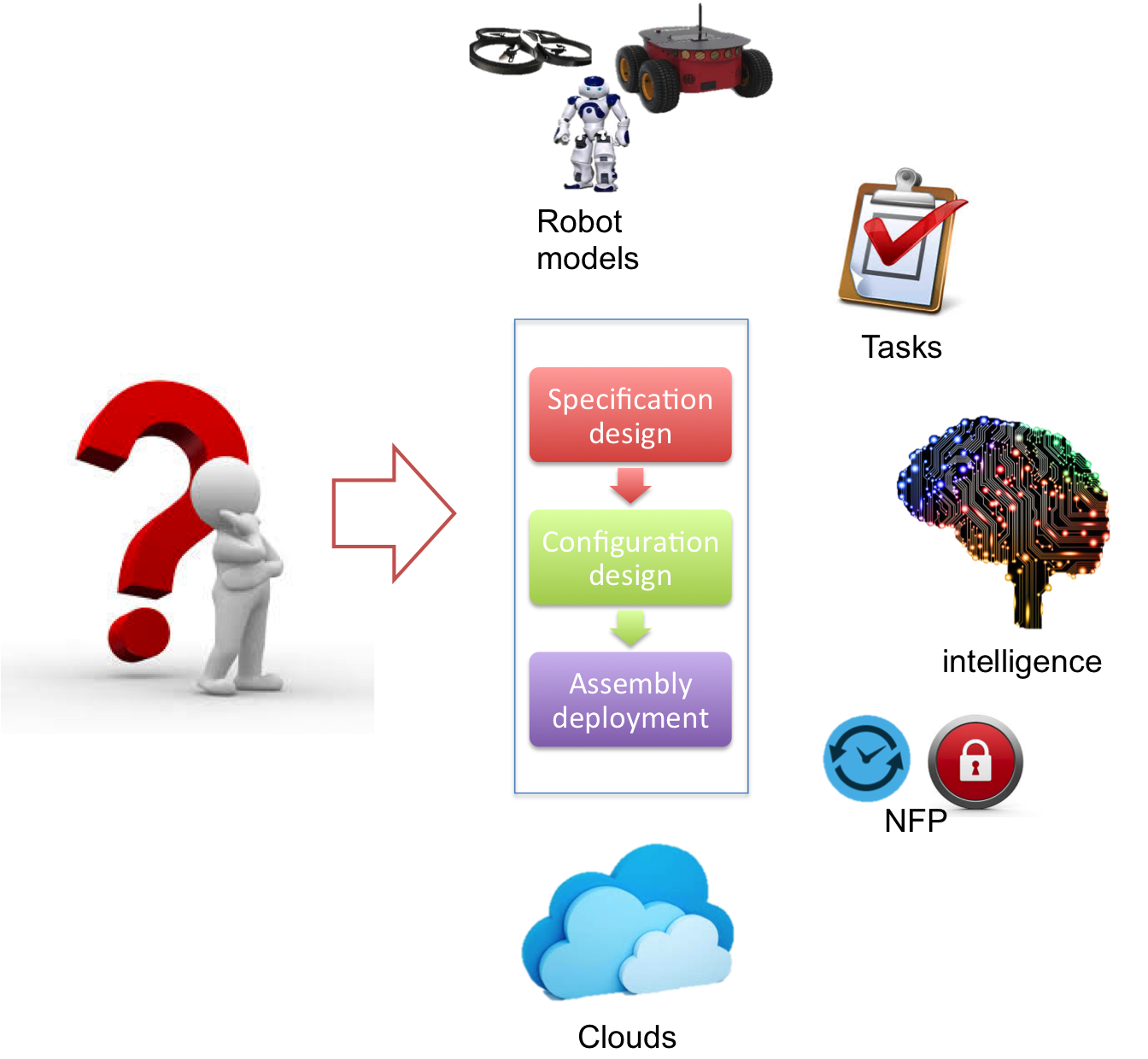}
\caption{Cloud robotic system architecture design}
\label{fig_global}
\end{figure}

During the process, five factors affect architecture design decisions \textit{robot models, tasks, intelligence, non-functional properties}, and \textit{Clouds}, as shown in Figure~\ref{fig_global}. 

\begin{itemize}
\item \textit{Robot model} describes the hardware model of the robots consisting of sensors and actuators etc.
\item \textit{Task} is the objective realized by robots.
\item \textit{Intelligence distribution} defines how to distribute the intelligence to robots and Clouds.
\item \textit{Non-functional properties} are non-functional requirements required to be exposed by Cloud robotic systems, such as security, realtime, safety etc.
\item \textit{Clouds} represent Cloud infrastructures (IaaS) used to deploy robotics services. Clouds can be mono-cloud or multi-Cloud.
\end{itemize}

\section{CRALA: A Domain Specific Language}\label{sec:dsl}

CRALA is a domain specific language for architecture-centric Cloud robotics, and it is also an architecture description language.   
CRALA models architectures at three separate abstraction levels, each designed in a different development phase as shown in Fig.~\ref{fig_process}. For now, the first version of CRALA presented in this paper mainly focuses on modeling essential elements of architectures and their basic properties, as the design concept of CRALA is to auto-develop and enrich the language by experimentation and real use cases. 
The three levels are as follows: 
%%%%%%%%%%%%%%%
\begin{enumerate}
\item \textit{Specification} defines the abstract architecture specification. It defines which
  functionality should be supplied by robotic systems. All the constituents of this architectural models are abstract and without any consideration of Cloud etc. 
\item \textit{Configuration} defines the sets of component or service implementations (classes) by searching and selecting from the component/service repository and defines how to group services and components in different virtual or physical machines by consideration of system requirements. 
\item \textit{Assembly} depicts how configuration is deployed on Clouds. This architecture model exactly depicts the current state of Cloud robotic system on Cloud.
\end{enumerate}

Table \ref{tab:term} presents the design decisions that should be made in each architecture level.
\begin{table}[t]
\renewcommand{\arraystretch}{1.3}
\caption{Decision decisions made in each architecture abstraction model}
\label{tab:term}
\centering
\newcolumntype{L}[1]{>{\hsize=#1\hsize\raggedright\arraybackslash}X}%
\newcolumntype{R}[1]{>{\hsize=#1\hsize\raggedleft\arraybackslash}X}%
\begin{tabularx}
{0.48\textwidth}{| L{0.8} | L{1.2} |}
\hline
Architecture & Defined Aspects\\
\hline
Architecture specification & 1) Functionalities of the system, 2) system non-functional properties\\
\hline
Architecture configuration & 
1) Component/service selection (for reuse) or implementation (for from scratch),
2) Component/service group, and
3) Operating system selection
\\
\hline
System assembly & 1) Cloud deployment, 2) Running state\\
\hline
\end{tabularx}
\end{table}

\subsection{Architecture Specification}
\begin{figure}[t]
\centering
\includegraphics[width=0.48\textwidth]{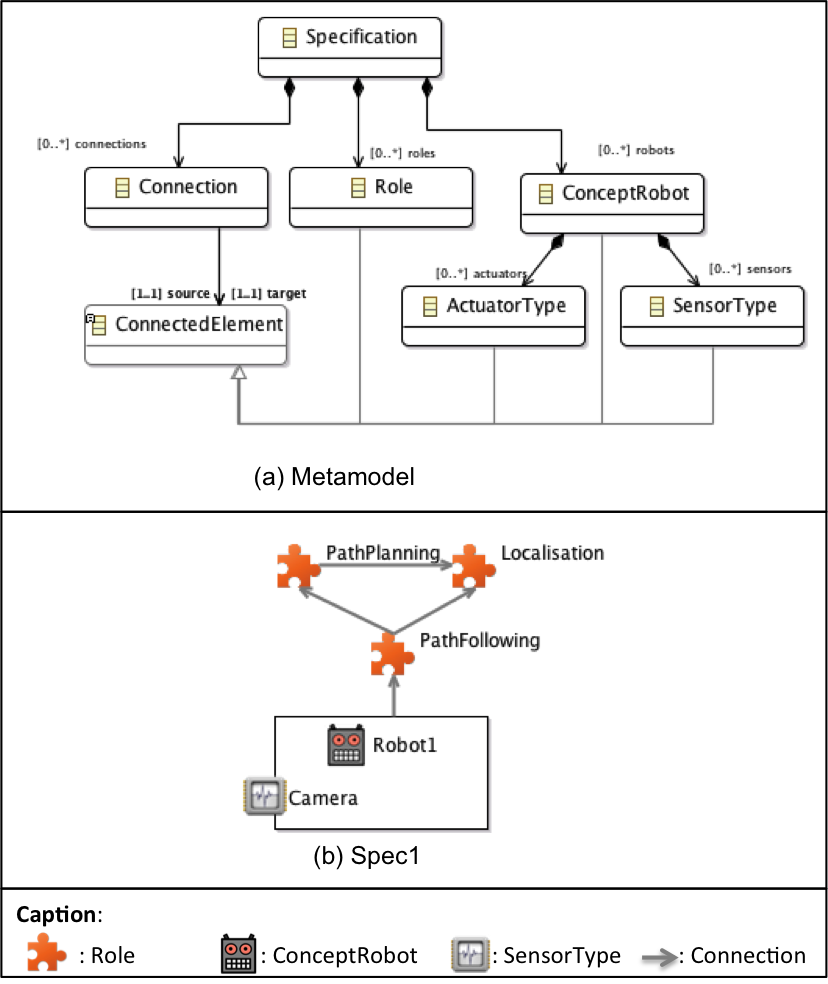}\\
\caption{Architecture specification Metamodel and example}
\label{fig_spec}
\end{figure}

Architecture specification is composed by \textit{component roles}, \textit{connections} and \textit{Concept robots}. 
The metamodel\footnote{In this paper, we ignore interfaces aspects and all attributes in metamodel for sake of simplicity.} of specification is illustrated in Fig.~\ref{fig_spec}(a).

\begin{itemize}
\item Component roles describe the roles that components should play in the system. In Cloud robotic systems, a roles could be a function (such as algorithm), a database, or a driver etc. A component role lists the minimum list
of interfaces (both required and provided) the component/service (will be selected or implemented in configuration level) should expose. On the one had, as they define the requirements of the architect (its ideal view) to guide the search for corresponding concrete
components (or service) in component (or service) repository, component roles are abstract and partial component (or service) representations. On the other hand, they can be used as the design specification for implementing new components or services from scratch. For example in Fig. \ref{fig_spec}(b), \textit{Spec1} defines three component roles to fulfill three different functionalities. 
\item Concept robots define the robots that will be included in the system with certain sensors or actuators to realize the functionalities of the system. At this level, concept robot is totally abstract, and it only defines the types of sensor and actuators. In specification, we do not precise the model of robot used. As shown in Fig. \ref{fig_spec}(b), \textit{Robot1} could be any robot with an camera, such as pionner, NAO etc.
\item Connections\footnote{For sake's simplicity, the details of connection and its constraints are not discussed in this paper.} define the communication between architecture elements including roles, robots, actuators and sensors. With CRALA connection constraints, the communication allowed could be categorized in three types: (1) the communication between component roles, (2) the communication between component roles (drivers) and sensors and (3) the communication between component roles (drivers) and actuators. However at specification level, we define also one kind of "abstract" connection between robots and component roles. The connection signifies the connected component roles must communicate with robots at next configuration level (components could locate directly in robots or connect with robots from Cloud.). 
\end{itemize}

\subsection{Architecture Configuration}
Architecture configurations are the second level of system architecture
descriptions. They result from the search and selection of real component classes (or web services) in a component (or service) repository. The metamodel is shown in Fig.~\ref{fig_config}.

\begin{enumerate}
\item  We precise which robots (\textit{RobotModel}) will be used in configuration. Robot models should expose all sensors and actuators specified in concept robot of \textit{Specification}. 
\item Component roles will be implemented by component classes or web services by selection or implementations according to different system requirements or used robot models. 

\begin{figure}[t]
\centering
\includegraphics[width=0.49\textwidth]{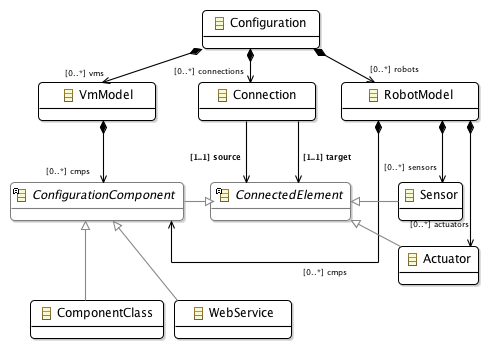}
\caption{Architecture Configuration Metamodel}
\label{fig_config}
\end{figure}

\begin{figure}[t]
\centering
\includegraphics[width=0.5\textwidth]{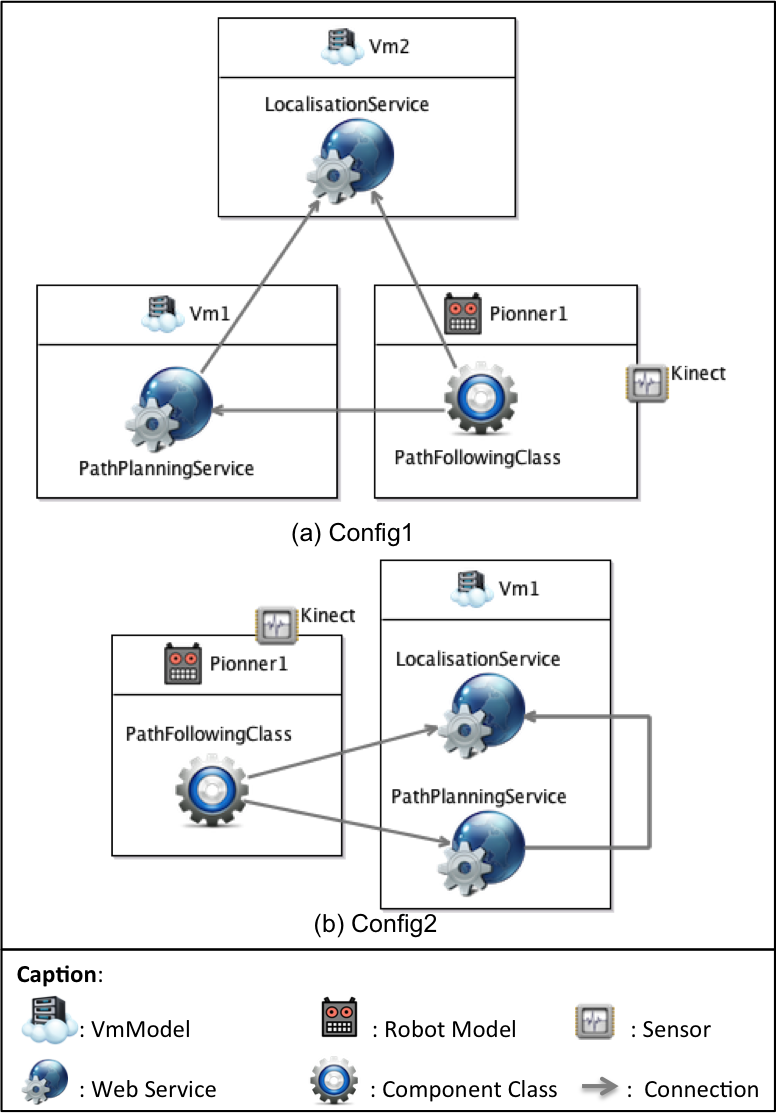}
\caption{Architecture Configuration Example \textit{Config1} and \textit{Config2}}
\label{fig_config_ex1}
\end{figure}

\begin{itemize}
\item Component class: A component class often can be characterized as: attributes, component interface, behaviors and properties.
\item Web service: A more formal and extended definition is the one offered by the W3C Web Services
working group\cite{w3cweb}:A Web service is a software system designed to support interoperable machine-to-machine interaction over a network. It has an interface described in a machine-processable format (specifically WSDL). Other systems interact with the Web service in a manner prescribed by its description using SOAP-messages, typically conveyed using HTTP with an XML serialization in conjunction with other Web-related standards. 
\end{itemize}

\item Then, components and services will be placed in robots or virtual machines\footnote{In the first version of CRALA, we ignore the possibility that besides virtual machines, for web services could also be placed in physical machines directly.}. 

\item Lastly, the connections should be established, for example which kind of communication protocols will be used in different situations. We could find the connection between robots and components are not permitted at this level (the same for next assembly level).
\end{enumerate}

According to the selection of different component classes and services and the distribution of these components in robots and VMs, it could lead to different configuration architectures, which implement the same specification architecture. 

Figure~\ref{fig_config_ex1}(a) and \ref{fig_config_ex1}(b) represent two different possible configuration architectures of specification \textit{Spec1} in Fig.~\ref{fig_spec}(a). In \textit{Config1}, two services \textit{LocalisationService} and \textit{PathPlanningService} locate in two separated VMs, and in \textit{Config2}, they are located in the same VM. The reliability of \textit{Config1} is better than \textit{Config2}, as if the VM of \textit{Config2} is broken, both two services are lost at the same time.
However, in \textit{Config1}, two services use more calculating resource, as they are located in two VMs compared to one.

\subsection{System assembly}
System assemblies are the third level of system
architecture descriptions. At the macroscopic level, They result from the deployment of VMs of configuration in Clouds. At the microscopic level, they result from the instantiation of the component
classes and the deployment of the web services from a configuration. The important thing is that they should provide a description of runtime software systems including Cloud deployment description. 

\begin{figure}[t]
\centering
\includegraphics[width=0.5\textwidth]{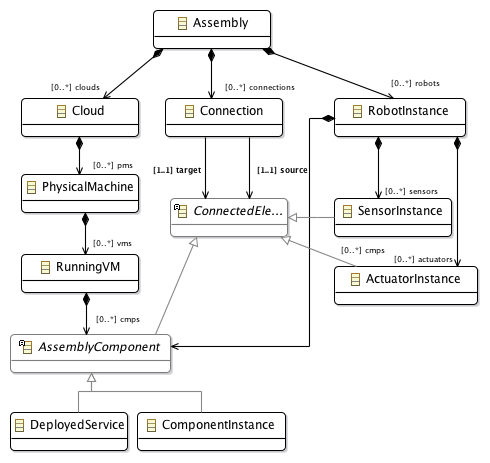}
\caption{Assembly Metamodel}
\label{fig_ass}
\end{figure}

\begin{figure*}[t]
\centering
\includegraphics[width=0.8\textwidth]{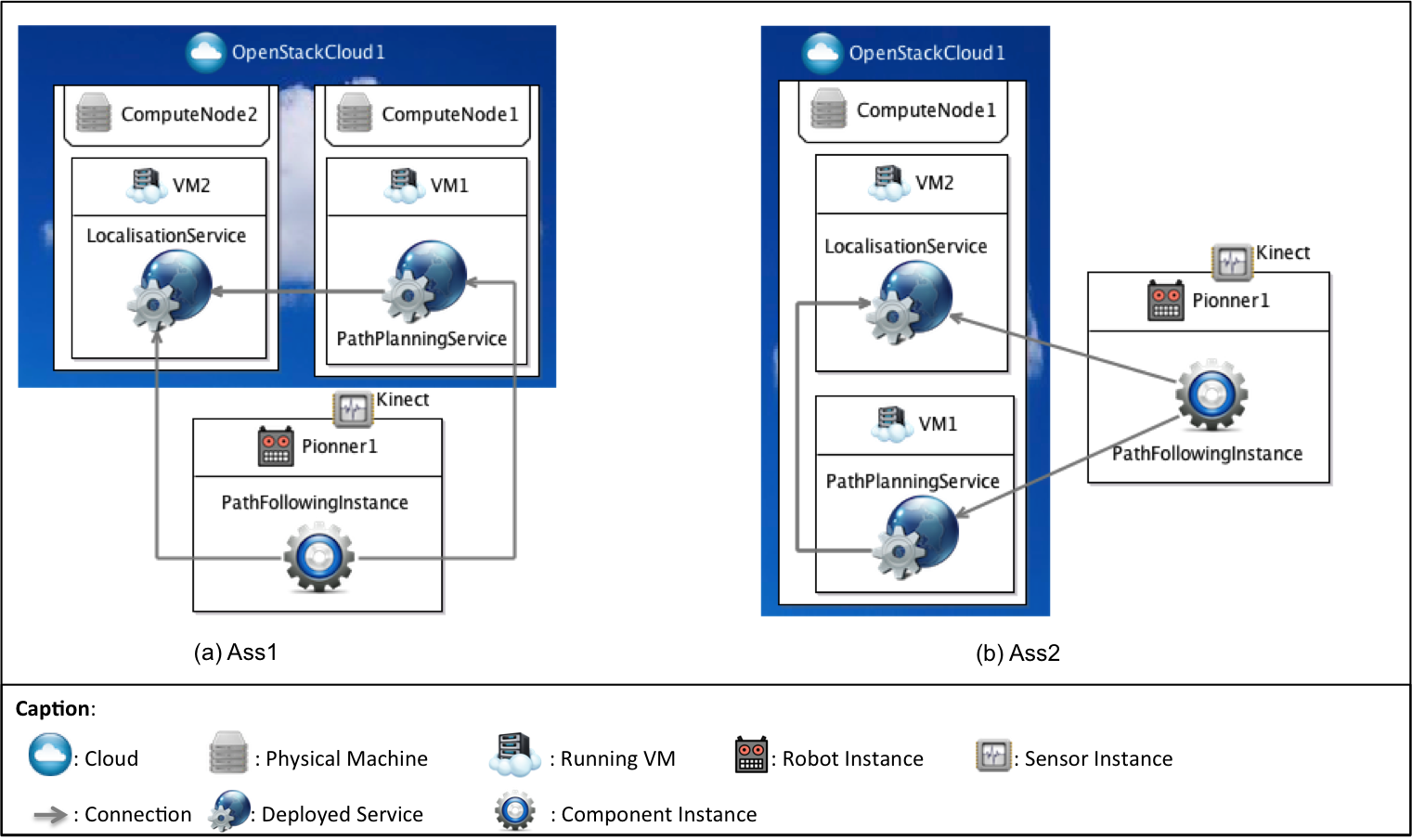}
\caption{Assembly Metamodel and examples}
\label{fig_ass_ex}
\end{figure*}

The metamodel of Cloud is illustrated in Fig.~\ref{fig_ass}. Each component or service are defined clearly which VMs or robots they are deployed and each VM is illustrated with which physical machine of which Cloud it is deployed. 

System assemblies are the most related level to Cloud. 
\begin{itemize}
\item Cloud: Different Clouds directly affect the deployment results of configurations. 
\begin{itemize}
\item Network: For example, if Cloud is an OpenStack~\cite{openstack} nova-network (FLAT) Cloud. The network of this kind of Cloud is linux-bridge, so all the VMs locate in the same network. If we want two VMs located in two subnets, it's impossible. However, with OpenStack neutron (SDN: Software-Defined Network) Cloud, it's possible. 
\item Scheduling: Scheduling in Clouds is complicated and it's extremely important for Cloud robotic systems. Some NFPs such as reliability, security could be directly applied by using different scheduling algorithms. One Cloud could apply multiple scheduling algorithms with different priorities. Normally different Clouds use different scheduling algorithms according to different requirements. According to different scheduling algorithms of Cloud, architecture configuration could be deployed in different ways, as shown in Fig.~\ref{fig_ass_ex}(a) and \ref{fig_ass_ex}(b). For the first example, two VMs are located in different physical machines, and for the second example, two VMs are located in the same physical machine. In \textit{Ass2}, two VMs communicate faster than \textit{Ass1}, as they locate in the same physical machine. However, in \textit{Ass1}, two VMs could profit the maximize RAM, as they are the only VM in each physical machine.
\end{itemize}
\item Physical machine: Normally in one Cloud, clients (tenants) could not see which physical machines locate their VMs. Only administrators could know this kind of information for security. In order to raise the clarity of Cloud, we add this information to assembly level. This could make easier to control Cloud robotic systems.
\end{itemize}

\subsection{The variability}
The variability of software architecture often cites SPL (Software Product Line). In SPL, the variability is inside in configuration level. A reference architecture could be implemented by different possible configurations with certain limit choices. In CRALA, variability is horizontal, which is reflected in the relationships between three levels. We use illustrating examples to explain how CRALA manages the variability of Cloud robotic systems from microscopic and macroscopic views. 

\begin{figure}[t]
\centering
\includegraphics[width=0.42\textwidth]{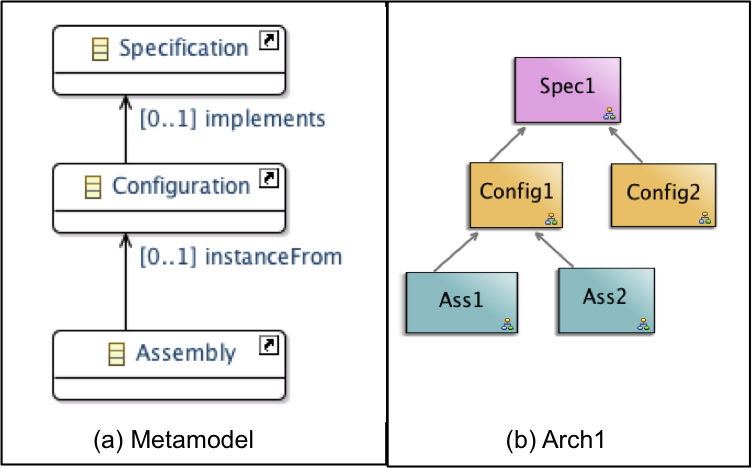}
\caption{The variability of example architectures}
\label{fig_ex}
\end{figure}

Firstly, from macroscopic view, the variability of architectures could be captured by their relationships between different architecture levels, as shown in Fig.~\ref{fig_ex} (a). Figure~\ref{fig_ex}(b) illustrates the relationships of example architectures \textit{Arch1} presented earlier in this section and it's generated automatically by CRALA toolsuite according to the relationships defined in architecture models (Fig.~\ref{fig_spec}(b), \ref{fig_config_ex1}(a,b) and \ref{fig_ass_ex}(a,b)). 

\begin{figure}[t]
\centering
\includegraphics[width=0.48\textwidth]{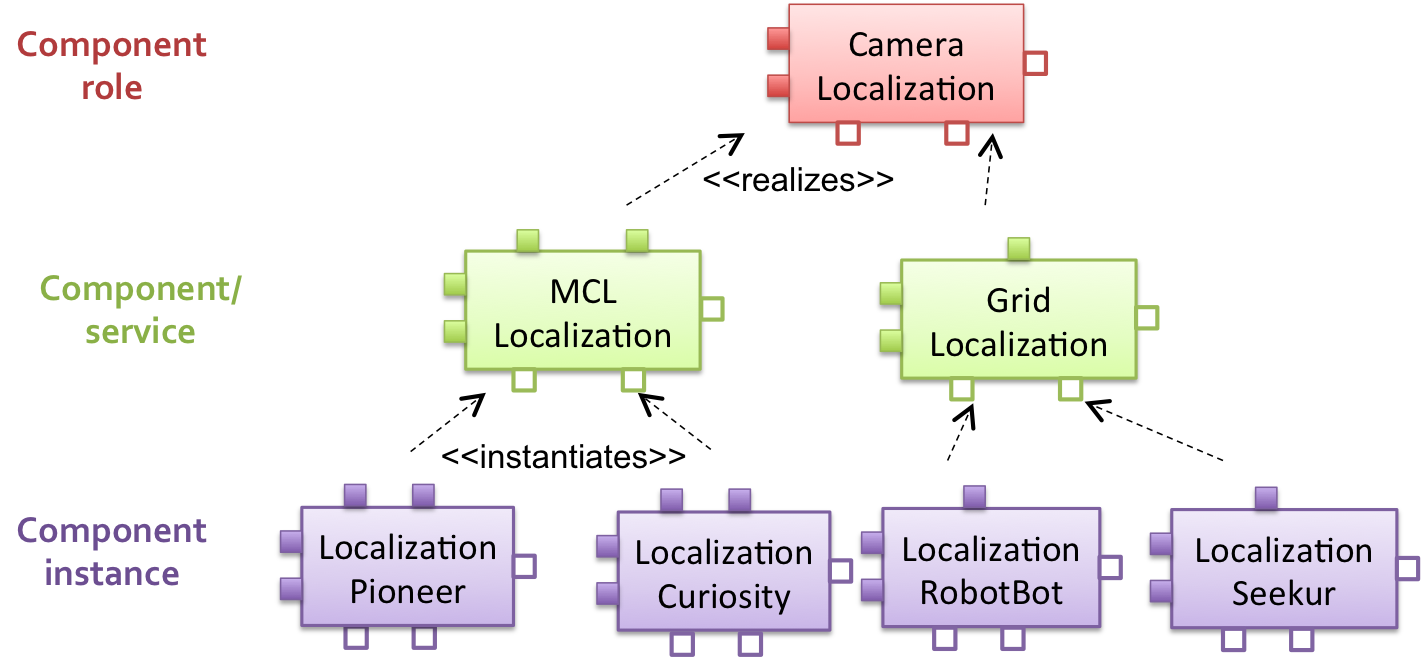}
\caption{The Localization component role, some possible concrete realizations and some of their instantiations}
\label{fig_cmex}
\end{figure}

Secondly from microscopic view, the variability is reflected by components. Fig.~\ref{fig_cmex} presents an example of components in three levels.
It shows the relationship of different component forms in three levels: component roles (specification), component/service (configuration) and component or service instance (assembly). 

In our viewpoint, if we could combine these two kinds of variability: horizontal and vertical in our future work, it will greatly increase the feasibility and reusability of CRALA.

\section{Implementation}
\begin{figure}[!h]
\centering
\includegraphics[width=0.4\textwidth]{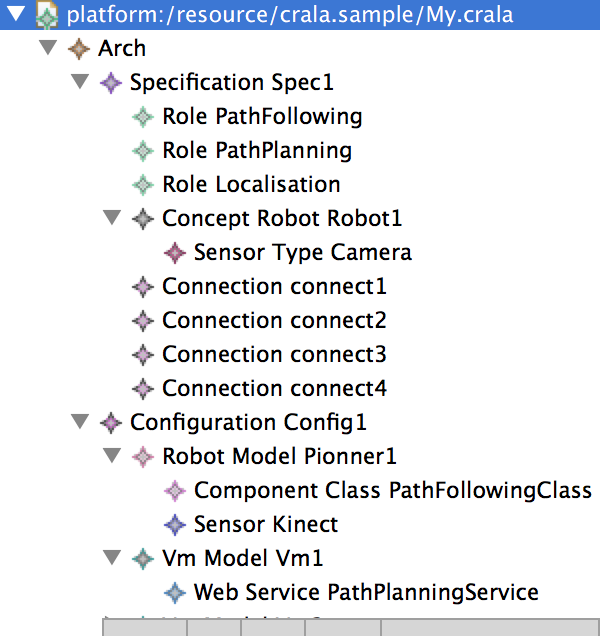}
\caption{A sample of CRALA Ecore editor}
\label{fig_ecore}
\end{figure}

We used the Ecore framework~\cite{Steinberg2008} and Sirius~\cite{Viyovic2014} for CRALA. Ecore allows create a tree editor for a DSL according to its metamodel, as show in Fig.~\ref{fig_ecore}. Sirius is an Eclipse project which allows you to easily create your own graphical modeling workbench including generating graph of models or editing graphical models. The models created by CRALA Ecore plugins could be automatically expressed in graphs. Figures ~\ref{fig_spec}(b), \ref{fig_config_ex1}(a,b), \ref{fig_ass}(a,b) and \ref{fig_ex} are graphs generated from CRALA models.  

\section{Conclusion and Future Work}
In this paper we investigate architecture design process for Cloud robotic systems and propose a domain-specific architecture description language for architecture-centric Cloud robotics. We present CRALA for describing Cloud robotic architectures, and show that linking architecture descriptions with Cloud deployment aspect allows mastering and controlling Cloud robotic systems and their variability. 
The proposed language is implemented by EMF and Sirius and we use a use case to illustrate CRALA. 

In future work, we aim to extend CRALA in several ways. We would like to develop some mechanisms to support the automatically developing process of architecture-centric Cloud robotic systems. First of all, how to search the correspondent and appropriate components or services in repository to construct architecture configuration automatically. Secondly, how to deploy the configuration on Cloud automatically. Then how to reorganize the system on Clouds when service failure. Our overall goal is to construct an intelligent development environment to construct Cloud robotic systems.

\bibliographystyle{IEEEtran}

%\bibliography{cloudrobotics}

\end{document}